\documentclass[sigconf]{acmart}

\usepackage{subcaption}
\usepackage{multirow}
\usepackage{paralist}
\usepackage{booktabs}

\usepackage[disable]{todonotes}
\usepackage{todonotes}
\usepackage{hyperref}
\usepackage{cleveref}

\hypersetup{draft}

\newcommand{\smalltt}[1]{{\small \texttt{ #1}}}

\def\BibTeX{{\rm B\kern-.05em{\sc i\kern-.025em b}\kern-.08emT\kern-.1667em\lower.7ex\hbox{E}\kern-.125emX}}
    
\copyrightyear{2019}
\acmYear{2019}
\setcopyright{acmlicensed}
\acmConference[ACM-BCB'19] {10th ACM International Conference on Bioinformatics, Computational Biology and Health Informatics}{September 7--10, 2019}{Niagara Falls, NY, USA}
\acmBooktitle{10th ACM Int'l Conference on Bioinformatics, Computational Biology and Health Informatics (ACM-BCB '19), September 7--10, 2019, Niagara Falls, NY, USA}
\acmPrice{15.00}
\acmDOI{10.1145/3307339.3342161}
\acmISBN{978-1-4503-6666-3/19/09}

\begin{document}
\fancyhead{}

\title[Drug-Drug Interaction Prediction Based on KG Embeddings and Conv-LSTM Network]{Drug-Drug Interaction Prediction Based on Knowledge Graph Embeddings and Convolutional-LSTM Network}

\author{Md. Rezaul Karim}
\affiliation{
    \institution{Fraunhofer FIT, Aachen, Germany}
    \institution{RWTH Aachen University, Germany}
}
\author{Michael Cochez}
\affiliation{
    \institution{Fraunhofer FIT, Aachen, Germany}
    \institution{RWTH Aachen University, Germany}
}
\author{Joao Bosco Jares}
\affiliation{
    \institution{Fraunhofer FIT, Aachen, Germany}
     \institution{RWTH Aachen University, Germany}
}
\author{Mamtaz Uddin}
\affiliation{
    \institution{Clinical Pharmacy and Pharmacology}
    \institution{University of Dhaka, Bangladesh}
}
\author{Oya Beyan}
\affiliation{
    \institution{RWTH Aachen University, Germany}
    \institution{Fraunhofer FIT, Aachen, Germany}
}
\author{Stefan Decker}
\affiliation{
    \institution{Fraunhofer FIT, Aachen, Germany}
    \institution{RWTH Aachen University, Germany}
    }

%
\renewcommand{\shortauthors}{Karim, Cochez, et al.}

%
\begin{abstract}
\todo[inline]{@MC There are problems with the hyperef in the document, that is why it is currently set to draft. This should be turned off for the final version. Hopefully the problems will solve themselves.}
\todo[inline]{@MRK The bibliography still needs several entries completed. Aim for no warnings left.}
\todo[inline]{@MC @MRK Find a name for the approach? }
Interference between pharmacological substances can cause serious medical injuries.
Correctly predicting so-called drug-drug interactions~(DDI) does not only reduce these cases but can also result in a reduction of drug development cost. 
Presently, most drug-related knowledge is the result of clinical evaluations and post marketing surveillance; resulting in a limited amount of information.
Existing data-driven prediction approaches for DDIs typically rely on a single source of information, while using information from multiple sources would help improve predictions.
Machine learning~(ML) techniques are used, but the techniques are often unable to deal with skew in the data.
Hence, we propose a new ML approach for predicting DDIs based on multiple data sources. 
For this task we use 12,000 drug features from DrugBank, PharmGKB, and KEGG drugs, which are integrated using Knowledge Graphs~(KGs).
To train our prediction model, we first embed the nodes in the graph using various embedding approaches. 
We found that the best performing combination was a
 ComplEx embedding method creating using PyTorch-BigGraph~(PBG) with a Convolutional-LSTM network and classic machine learning based prediction models. 
The model averaging ensemble method of three best classifiers 
yields up to 0.94, 0.92, 0.80 for AUPR, F1-score, and MCC, respectively during 5-fold cross-validation tests.
\end{abstract}



\maketitle

\section{Introduction}
Drug-drug interactions~(DDIs) are (often) preventable causes of medical injuries that occur when a drug causes a pharmacokinetic~(PK) or pharmacodynamic~(PD) effect on the body when it is taken together with another drug~\cite{wang2017predicting,takeda2017predicting}. 
They are a common cause of adverse drug reactions~(ADRs) and increased healthcare costs~\cite{cheng2014machine}. 
The majority of ADRs are caused by unintended DDIs and occasionally arise through co-prescription of drugs. 
While it would be ideal for identifying all possible DDIs during clinical trials, interactions are frequently reported after the drugs are approved for clinical use.
ADRs are a significant threat to public health, as shown in a study by Shtar et al. 
They found that about 6.7\% of hospital readmission occurred because of ADRs with a fatality rate of 0.32\% in the USA in 2014. 
In that year, as many as 807,270 cases of serious ADRs were reported in the United States, resulting in 123,927 lost lives~\cite{shtar2019detecting}.

For example, acetylsalicylic acid, commonly known as aspirin, is a drug used for the treatment of pain and fever due to various causes. This medicine has both anti-inflammatory and antipyretic effects, which inhibits platelet aggregation and is used in the prevention of blood clots and myocardial infarction. 
However, the risk or severity of hypertension can be increased (e.g., negative drug-drug interaction) when acetylsalicylic acid is combined with 1-benzylimidazole~\cite{drugbank5}.
Predicting potential DDIs reduces unanticipated drug interactions, lowers drug development costs, and can be used to optimize the drug design process. 
Thus, the study of DDIs and ADRs is important in both drug development and clinical application, especially for co-administered medications.
Since the majority of ADRs occur between pairs of drugs, they have become the focus of research and clinical studies\cite{wang2017predicting}. 
To further reduce costs and to make the analysis of large amounts of interactions possible, automated methods for identifying ADRs are needed.
Current approaches involve clinical evaluation of drugs and post-marketing surveillance.
Here, features are extracted from drug properties such as targets, side-effects, chemical properties, fingerprint, and drug indications.
Then statistical methods and various supervised ML algorithms 
(like, e.g., decision tree~(DT), Naive Bayes~(NB), k-nearest neighbors~(k-NN), logistic regression~(LR), support vector machine~(SVM), random forest~(RF), and gradient boosting trees~(GBT))
~\cite{abdelaziz2017large} are employed.

Deep learning-based approaches, which can utilize deep features, are mostly unexplored in the context of DDIs prediction. 
While a deep architecture like a convolutional neural network~(CNN) is good at reducing frequency variations by acting as a feature extractor~\cite{zhang2018hate}, a long short-term memory~(LSTM) network is good at temporal modeling and learning orderly sequences from a large feature space~\cite{Conv_LSTM1}. 
By combining these two deep architectures, the convolutional-LSTM(\texttt{Conv-LSTM}) can capture both locally and globally important drug features which we found to lead to more accurate DDI predictions~\cite{Conv_LSTM1}. 
However, the features which have been traditionally used for these approaches form either a large and sparse binary matrix or a dense, but small similarity matrix, making them not ideal for training ML models~\cite{celebi2018evaluation}. 
Further, an increasing amount of drug and small molecules data is being generated, and state-of-the-art approaches still rely on the analysis on a limited number of data sources only, e.g. DrugBank.

To incorporate multiple data sources, Knowledge Graphs are a powerful tool, and many biomedical knowledge bases have been published in this form.
In this graph, the nodes represent different entities like drugs, diseases, protein targets, substructures, side effects, and pathways.
See e.g.,~\cite{wang2017predicting,celebi2018evaluation} for examples on using Knoweldge Graphs for DDI prediction.
Once the data is in the form of a Knowledge Graph, we have to extract information from it as features for our interaction predictors.
To do this, we use embedding methods which project each node in the graph to a dense vector.
%
%
%
%
%
%
%


In our work, we consider more sources of DDIs as others.
Scientific literature that has predicted DDIs very accurately is often ignored as ground truth in related work. 
In this paper, we collected DDI information from 
\begin{inparaenum}[]
	\item DrugBank~\cite{drugbank5}, 
	\item the Kyoto Encyclopedia of Genes \& Genomes~(KEGG)~\cite{KEGG}, \item TWOSIDES~\cite{tatonetti2012data}, and 
	\item scientific literature
\end{inparaenum}.
Then, we created an integrated KG using data from DrugBank, KEGG drug, PharmGKB~\cite{pharmgkb}, and OFFSIDES~\cite{tatonetti2012data} (excluding data already in the above mentioned DDI data).
To transform the information from this graph in a format suitable for the prediction models, we applied different KG embedding techniques.
Then, we trained several ML models as baselines and also performed experiments with the Conv-LSTM model. 
The key contributions of this paper can be summarized as follows:  

\begin{itemize}
\item We have created a dataset with 2,898,937 drug-drug interaction pairs; we believe that this is the largest available.
\item We have prepared a large-scale integrated KG about DDIs with data from DrugBank, KEGG, OFFSIDES, and PharmGKB having 1.2 billion triples.
\item We have evaluated different KG embeddings techniques with different settings to train and evaluate ML models.
\item We provide a comprehensive evaluation with details analysis of the outcome and comparison with the state-of-the-art approaches and baseline models. 
\item We found that a combined CNN and LSTM network called \texttt{Conv-LSTM} for predicting DDIs leads to the highest accuracy.
\end{itemize}


This paper is structured as follows: \cref{rw} discusses related works with their emerging use cases and potential limitations. 
\Cref{mm} details the proposed approach, including problem formulation, data collections, KG construction, graph embeddings, network constructions, and training. 
The results of our experiments can be found in \Cref{results}, where we also discuss the key findings of the study. \Cref{conclusion} provides some explanations of the importance, highlights the limitations of the study reported, and discusses some future works before concluding the paper.
To avoid confusion, we use the terms node and drug interchangeably throughout the paper.

\section{Related Work}
\label{rw}
Till date, DDI prediction is a non-trivial research problem in pharmacology, and numerous approaches have been proposed to predict novel DDIs by employing various data sources.
Traditional work relies on in vitro and in vivo experiments and focuses on small sets of specific drug pairs and had laboratory limitations~\cite{duke2012literature}. 
With the emergence of available biomedical data, researchers moved the focus towards automatically populating and completing biomedical KGs using large-scale structured databases and text publicly available~\cite{celebi2018evaluation}. 
In this scope, the Bio2RDF project made 35 life sciences datasets as linked open data~(LOD) in RDF, in which similar entities are mapped in different KGs and built large heterogeneous graphs that also contain biomedical drug-related facts. 
Although these approaches made available numerous biomedical KGs, they often contain incomplete and inaccurate data that impede their application in the field of safe medicine development~\cite{celebi2018evaluation}. 
Lately, ML and text mining based approaches were used in which pharmacological similarities of drugs as features are used by regarding the DDIs prediction task as a link prediction problem. 

Different drug similarity metrics are used for inferring DDIs and their associated recommendations in which LR is trained to calculate the maximum likelihood by using known DDIs~\cite{gottlieb2012indi}. 
Similarly, DDIs using phenotypic, chemical, biological, therapeutic, structural, and genomic similarities of drugs are used for predicting DDIs~\cite{cheng2014machine}. 
Other investigations used pharmacological and graph qualities between drugs~\cite{cami2013pharmacointeraction} or drug structural similarities and interaction networks incorporating PK and PD knowledge~\cite{takeda2017predicting} using LR. 
Peng et al.~\cite{li2015large} developed a Bayesian network
, which combines molecular drug similarity and drug side-effect similarity to predict the combined effect of drugs.
Lately, Andrej K. et al.~\cite{kastrin2018predicting} formulated the DDIs prediction problem as a binary classification problem to predict
unknown DDIs in 5 arbitrary databases such as DrugBank, KEGG, NDF-RT, SemMedDB, and Twosides.
In these works, supervised ML approaches such as DT, NB, k-NN, LR, SVM, RF, and GBT are mostly used for predicting DDIs from topological and semantic similarity features.
However, feature-based approaches only predict binary DDIs or those that have been pre-defined in structured databases and suffer from robustness caused by data sparsity and vast computation requirements.
Similarity-based approaches, in contrast, do not allow for the calculation of various similarities for many drugs due to lack of drug information~\cite{celebi2018evaluation}.

Several other approaches are proposed using biomedical KGs and text embedding in which graph embedding is utilized~\cite{wang2017predicting} to overcome issues such as data incompleteness and the sparsity problem. 
These learned embeddings are then mostly used for predicting DDIs.
Other works have utilized text mining~\cite{percha2012discovery,tari2010discovering,duke2012literature} to predict and evaluate new DDIs in which either drug-related data from scientific literature was discovered from large health information exchange repository~\cite{duke2012literature} or automated reasoning has been developed to derive new enzyme-based DDIs from MEDLINE abstracts~\cite{tari2010discovering}. 
With the abundance of biomedical data characterizing drugs and their associated targets, these methods cannot fuse multiple sources of information and perform inference over the network of drugs effectively.
Therefore, approaches employing KGs embeddings and ML-approaches~\cite{wang2017predicting,abdelaziz2017large, celebi2018evaluation,hallstedt2018drugdisease} have emerged.
In particular, approaches based on KG embeddings are powerful predictors and outperform state-of-the-art approaches for inferring new DDIs.
Most of the embedding methods are translation-based; embeddings are built by treating relations as translations from the head entity to tail entity.
In these methods, the vector embeddings are created such that $\mathbf{h}\bigoplus\mathbf{r} \approx \mathbf{t}$ where $(h,r,t)$ is a triple of knowledge base (i.e., the relation $r$ holds between $h$ and $t$).
For this equation, there are various options for the $\bigoplus$ operator (see e.g.,~\cite{FTE,TransE, TransD}).

With these ideas in mind, the DDIs prediction framework called Tiresias~\cite{abdelaziz2017large} is proposed in which various sources of drug-related data and knowledge are used as inputs and provides DDIs predictions as outputs using an LR classifier.
The process starts with semantic integration of data into a KG describing drug attributes and relationships with various related entities such as enzymes, chemical structures, and pathways.
The KG is then used to compute several similarity measures between the drugs in a scalable and distributed framework.
A recent approach called PRD~\cite{wang2017predicting} was proposed for predicting DDIs in which graph embedding techniques such as TransE, TransD, TransH, HolE are employed to overcome the data incompleteness and sparsity issues~\cite{abdelaziz2017large}.
First, a large-scale drug KG is created from different sources containing biomedical texts, which are then embedded into a common low dimensional space into a continuous vector space in which both entities and relations were considered. 
The learned embeddings are subsequently used to predict the DDIs using a rich DDI triple encoder~(RDTE) network in which an encoder incorporates the drug-related information to obtain the DDI relation representation. 
The decoder then reconstructs the embedding vector from the latent representation and is used to predict the labels for potential DDIs.

However, most translation-based embeddings are limited in their capacity to model complex and diverse objects, including important properties of relations, such as symmetric, transitive, one-many, many-to-one, and many-many relations in KGs~\cite{FTE}. 
As KG embeddings techniques show state-of-the-art performance, generating quality feature vectors using appropriate embedding methods plays a significant role.
A more recent work~\cite{celebi2018evaluation} employed KG embedding methods to extract feature vector representation of drugs using LOD to predict potential DDIs.
The effects of DDIs prediction accuracy using LR, NB, and RF is also investigated on a single source (the Bio2RDF DrugBank v4 dataset) with different embedding methods such as RDF2Vec, TransE, and TransD. 

In these approaches, DDIs information extraction from biomedical texts and drug event reports using text mining~(TM) and then inferring DDIs by integrating knowledge from several sources are two typical steps~\cite{tatonetti2012data,percha2013informatics,kastrin2018predicting}. 
Although, the way DDIs are extracted and the predictions made vary across methods. 
Numerous approaches have extracted DDIs from biomedical text using knowledge-rich and knowledge-poor features~\cite{DDIE1} or from lexical, syntactical, and semantic-based features~\cite{DDIE2}. 
Other approaches, focus on classifying DDIs in which an SVM is trained using drug features generated by similarity measures~\cite{DDIE3, DDIE4} or by exploiting linguistic information~\cite{DDIE5}.
Apart from these, other approaches have employed text mining for extracting DDIs from a semantically annotated corpus of documents describing DDIs from DrugBank and MEDLINE abstracts~\cite{DDICorpus1, DDICorpus2}.

\section{Materials and Methods}
\label{mm}
\begin{figure*}
	\centering
	\includegraphics[width=0.9\textwidth]{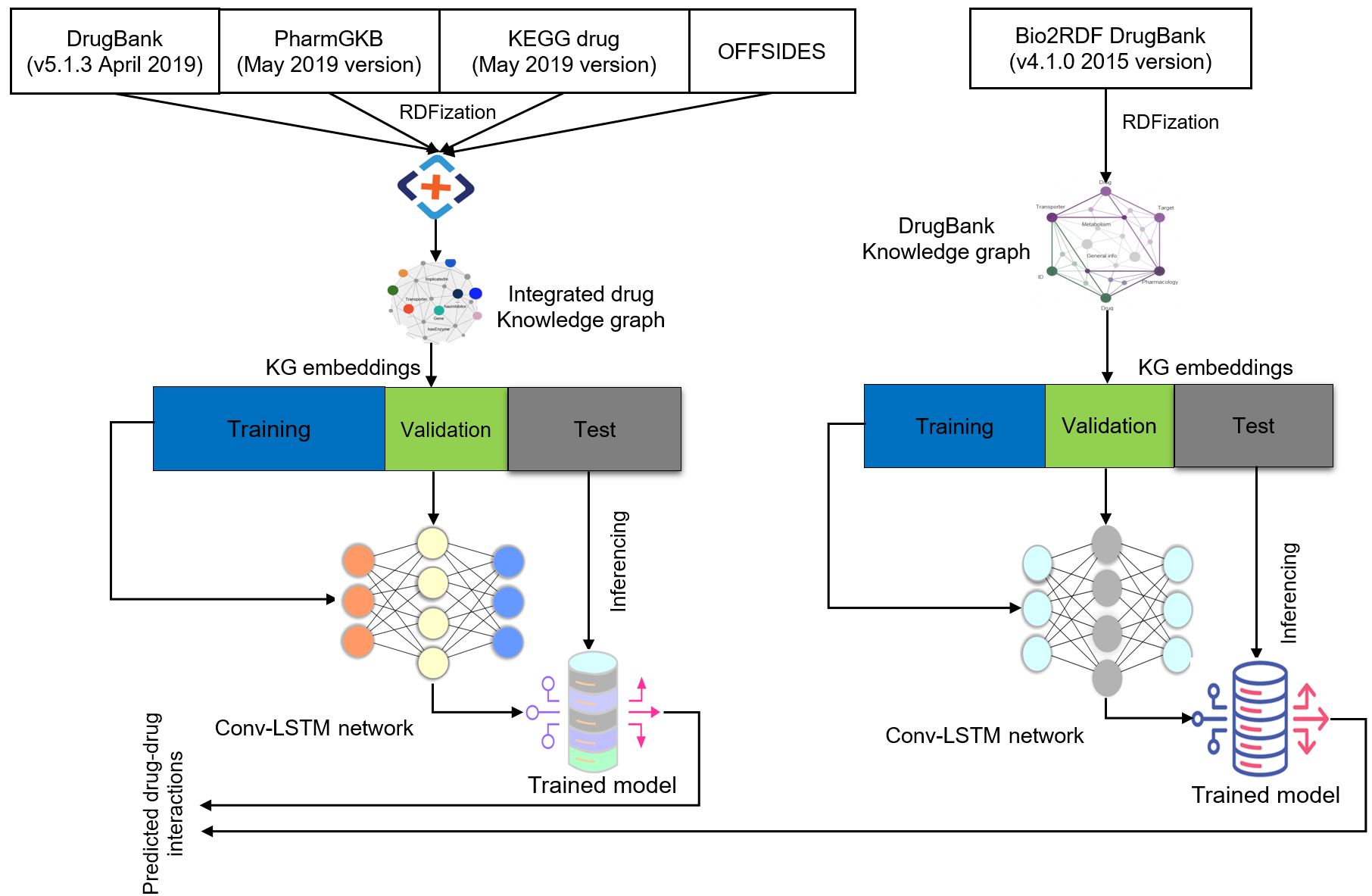}	
	\caption{workflow of the proposed approach for predicting DDIs}	
	\label{fig:workflow}
\end{figure*}

In this section, we discuss our methods in detail, including the problem formulation, data collection and integration, KG embeddings, the Conv-LSTM network construction, and the network training with hyperparameter optimization. 
The last step is inferencing DDI predictions. \Cref{fig:workflow} shows the workflow of the proposed approach. 

\subsection{Problem formulation}
Since DDIs form a complex network in which nodes refer to drugs and links refer to potential interactions, we approach the DDIs prediction task as a link prediction problem similar to Shtar et al.~\cite{shtar2019detecting}.
Given a directed DDI KG as $G=(V,E)$ in which each edge $e=(u,v) \in E$ represents an interaction between drugs $u$ and $v$. 
Let $N$ denotes the number of drugs, we can define the DDIs matrix $Y \in \{0,1\}^{N x N}$ as follows: 
 
\begin{equation}
    y_{u,v}=\left\{\begin{array}{ll}{1,} & {\text{if interaction exits between drugs u and v }} \\ {0,} & {\text{otherwise.}}\end{array}\right.  
    \label{eq:ddi_}
\end{equation}

In \cref{eq:ddi_}, a value of 1 for $y_{u,v}$ indicates an existing interaction between drugs $u$ and $v$. However, a value of 0 does not mean that an interaction does not exist in the KG, but it could be that the interaction has not yet been discovered~\cite{shtar2019detecting}.
Next, we proceed to DDIs extraction to be followed by KG construction. 

\subsection{DDIs extraction and KG construction}
Since creating an integrated KG and extracting DDIs are the two most important steps in our approach, we first focus on the data and knowledge source selection.
We constructed our integrated knowledge graph based on the drugs and drug-target related data from DrugBank, KEGG drug, and PharmGKB. 
On the other hand, OFFSIDES, TWOSIDES, and scientific literature from MEDLINE are used for finding DDI with enough evidence.  

\subsubsection{Data collection}
At the current time, there is no automated method or data source available which would provide complete DDI information.
Moreover, the available data is spread over multiple sources.
Therefore, we rely on several sources for collecting drug-related data.
The \emph{DrugBank} database is a bioinformatics and cheminformatics resource that combines detailed drug-related information, including chemical, pharmacological, and pharmaceutical data with comprehensive drug target information.

The \emph{PharmGKB} database\footnote{DrugBank v5.1.3, April 02, 2019} contains 12,664 drug entries including 2,588 approved small molecule drugs, 1,287 approved biotech drugs, 130 nutraceuticals and over 6,305 experimental drugs~\cite{drugbank5}.%
The \emph{PharmGKB} database is a repository for genomic, molecular, and cellular phenotype data.
It also contains clinical information and the impact of genetic variation on drug response about people who have participated in pharmacogenomics research studies. 
PharmGKB contains genes, diseases, drugs, and pathways related data as well as detailed information on 470 genetic variants affecting drug metabolism.

The \emph{KEGG} databases contain metabolic pathways that are hyperlinked to metabolite and protein/enzyme information.
As of May 2019, KEGG drug database has 10,979 drugs related information and 501,689 DDIs relations.
Finally, the \emph{OFFSIDES} database, which contains drug effects mined from adverse event reports based on PharmGKB~\cite{kastrin2018predicting,percha2013informatics}, reports 438,802 drug side-effects. 
From these sources, we create two datasets: i) the DDI dataset which contains drug-drug interaction pairs, ii) the knowledge graph which we will later use as background knowledge for interactions. The latter does not contain any explicit information about the interactions.

\subsubsection{DDI extraction}
We employ a semi-supervised approach for extracting DDIs from the sources mentioned above.
We parsed the DDI information from the provided XML file from DrugBank and compiled an edge list of drug identifier combinations, which gives us 2,641,889 pairwise DDIs and 2,630,796 unique DDIs spanning 12,112 drugs.
Although the KEGG drug database has 10,979 drugs related and 501,689 DDIs relations, mapping to DrugBank identifiers~(IDs) results in only 58,205 interactions because of missing mappings.

Data from TWOSIDES~\cite{Twosides}, which is a comprehensive source of polypharmacy ADRs is also used, but interactions are restricted to those that cannot be ascribed unambiguously to either drug alone.
Therefore, a collection of the drug pairs for the interacting compounds built in literature~\cite{kastrin2018predicting} is used in which the PubChem IDs are used to map TWOSIDES IDs to DrugBank IDs. 
This way, we obtained a list of 19,020 DDIs between 351 compounds and 63,473 distinct pairwise DDIs between 645 drugs. 

Dhami et al.~\cite{confusion} have identified that 
a few DDIs reported in the DrugBank dataset are less evident to interact with each other.
We relied on multi-source evidence for those drug pairs and removed from the DrugBank DDI list the contradictory ones.
Next, Zhang et al.~\cite{confusion4} reported 145,068 DDIs\footnote{\url{https://astro.temple.edu/~tua87106/ddi.html}} based on label propagation prediction using clinical side-effects.
We added these interactions to our dataset.
Finally, Sridhar et al.~\cite{sridhar2016probabilistic} listed top-ranked ten predictions for interactions unknown in DrugBank using their PSL model; also, these pairs were added to our dataset. 

\begin{figure}
	\centering
	\includegraphics[width=0.5\textwidth]{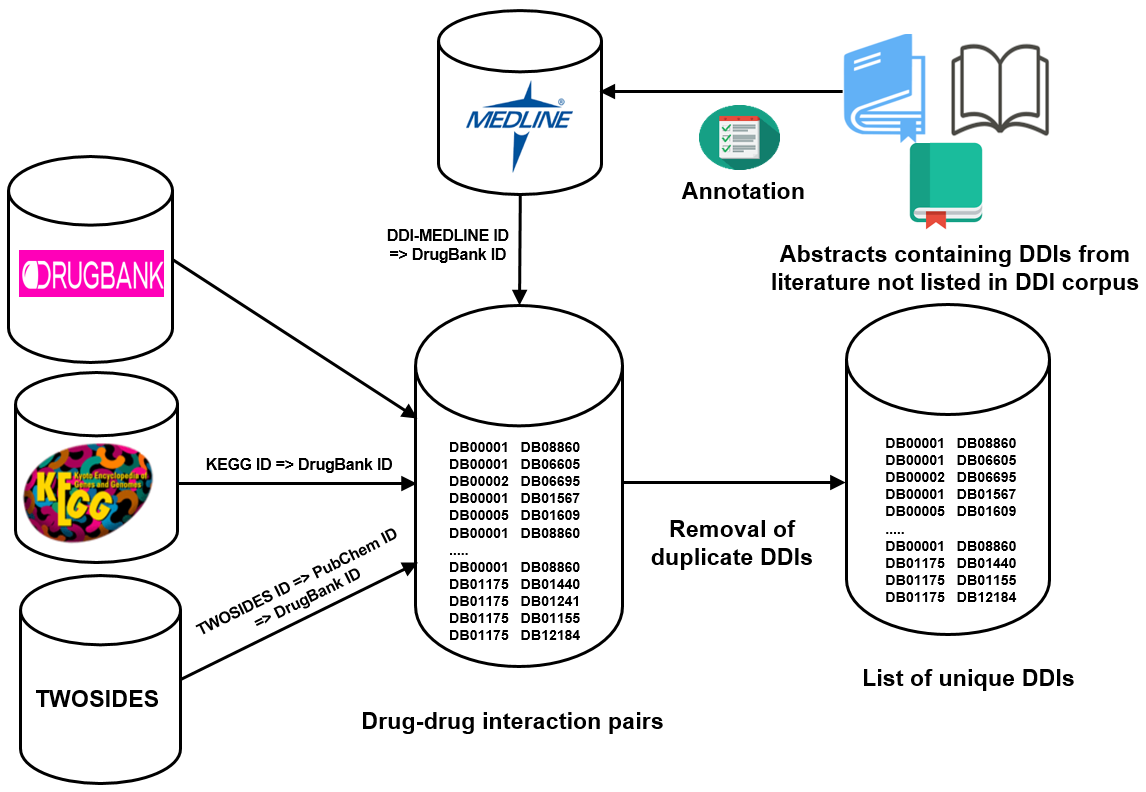}	
	\caption{extraction of DDIs from different scientific sources}	
	\label{fig:ddis}
\end{figure}

Besides these, we also incorporate the interactions from the 227 MEDLINE abstracts from the DDI corpus~\cite{DDICorpus1,DDICorpus2}. 
This contributed 327 DDIs based on 1,826 pharmacological substances.
Additionally, some abstracts are also used that are not listed in the DDI corpus e.g. \cite{confusion,sridhar2016probabilistic,confusion3,confusion4}.
for these, the annotation guidelines\footnote{\url{http://hulat.inf.uc3m.es/DrugDDI/annotation_guidelines_ddi_corpus.pdf}} developed by domain experts were used. 
A certified pharmacist verified these annotations.
The overall DDI dataset consists of information from all these sources combined. An overview of the process is shown in \cref{fig:ddis} and statistics are collected in~\cref{table:stat} where it should be noted that duplicates between the data sources are removed to obtain the final number of interactions (2,898,937). 

\begin{table}
	\caption{Statistics of drug-drug interactions}
	\label{table:stat}
	\centering
	\small
	\begin{tabular}{lr}
		\toprule
		\bf{Database/source}  & \bf{\#DD-interactions}  \\ 
		DrugBank &	2,630,796 \\
		KEGG Drug &	40,540 \\
		TWOSIDES &	82,493 \\
		MEDLINE~\cite{DDICorpus1}, other sources~\cite{confusion,sridhar2016probabilistic,confusion3,confusion4} &	145,108 \\
		\midrule
		\textbf{Total DDIs} & 2,898,937 \\
		\bottomrule
	\end{tabular}
\end{table}

\subsubsection{KGs construction and integration}
To create our integrated knowledge graph, we used data from DrugBank, KEGG, OFFSIDES, and PharmGKB.
Although PharmGKB does not contain DDI information, it publishes lexicons of known drug names and synonyms as well as gene and disease terms and data on genetic pathways. The dataset is directed at clinicians and researchers.
Previously, Bio2RDF~\cite{belleau2008bio2rdf} created a large RDF graph that interlinks data from major databases containing biological entities such as drugs, proteins, pathways, and diseases.
Using that data would have been an option, but the latest version~(i.e., v4.0) is already rather outdated.
Instead, we collected the raw DrugBank, KEGG drug, PharmGKB, and OFFSIDES data from the respective portals and converted them into RDF using a modified version of Bio2RDF scripts\footnote{\url{https://github.com/rezacsedu/DDI-prediction-KG-embeddings-Conv-LSTM/scripts}}.
Then each RDF KG was uploaded to a blazegraph RDF triplestore in named graph\footnote{\emph{Zenodo} download link: \url{https://zenodo.org/deposit/3270566}}.
Then similar to literature~\cite{wang2017predicting}, federated SPARQL queries are executed based on the `billion triples benchmark'~\cite{saleem2018largerdfbench} to extract selected triples. 
For our dataset five types of drug-related entities, namely drugs, genes, proteins, pathways and enzymes, and phenotype~(i.e., disease, side-effects), are included.
Further, nine types of biological relations are considered: 
\begin{inparaenum}
\smalltt{(drug, \textbf{hasTarget}, protein)}, 
\smalltt{(drug, \textbf{hasTarget}, gene)}, 
\smalltt{(drug, \textbf{hasEnzyme}, protein)}, 
\smalltt{(drug, \textbf{hasEnzyme}, gene)}, 
\smalltt{(drug, \textbf{hasTransporter}, protein)}, 
\smalltt{(drug, \textbf{hasTransporter}, gene\footnote{e.g. Polymorphisms in the ABC drug has transporter gene MDR1})}, 
\smalltt{(protein, \textbf{isPresentIn}, pathway)}, 
\smalltt{(gene, \textbf{isPresentIn}, pathway)}, and
\smalltt{(pathway, \textbf{isImplicatedIn}, phenotype)}.
\end{inparaenum}

Before the extraction, mappings are created based on \smalltt{owl:sameAs} and \smalltt{owl:equivalentProperty} axioms in which respective drug identifiers are mapped to DrugBank IDs as shown in \cref{fig:ddis}. 
Although genes contain the information needed to make functional molecules called proteins, we considered relation around genes and protein distinct, since PharmGKB contains information about both genes and proteins.
The extracted triples are formed the triples in our drug KG in the form (subject, predicate, object), indicating that the subject has the specified relation to the object.
Since this integrated knowledge graph should not contain any explicit information about drug-drug interactions, there is no information in the form of \smalltt{drugbank\_vocabulary:ddi-interactor-in} and \smalltt{kegg\_vocabulary:Interaction} from the DrugBank and KEGG drug KG, respectively.
The number of triples, entities, and relation types for the individual KGs and the integrated KG are given in \cref{table:stat2}. 
Next, we will elaborate on how we prepared this data as an input for our classifiers.

\begin{table}
	\caption{Statistics of the data sources}
	\label{table:stat2}
	\centering
	\small
	\begin{tabular}{lrrr}
		\toprule
		\textbf{Knowledge graph} & \bf{\#Triples}  & \bf{\#Entities}  &  \bf{\#Relation types} \\ 
		DrugBank & 7,740,864 &	2,116,569 &	72   \\
		KEGG & 308,690 &	107,916 &	41   \\
		OFFSIDES & 438,802 & 875,985 &	12   \\
		PharmGKB & 2,793,078 &	1,583,910 &	135   \\
		\midrule
		\textbf{Total} & 11,281,434 & 4,587,380 &	260   \\
		\bottomrule
	\end{tabular}
\end{table}

\subsection{Knowledge graph embeddings}
We used the information of our knowledge graph for predicting the interaction between each pair of drugs.
However, ML classifiers do typically expect their input as fixed-length vectors.
Hence, we perform a KG embedding procedure to encode the information from the graph into dense vectors.
KG embedding consists of three steps: representing entities and relations, defining a scoring function, and learning entity and relation representation~\cite{lin2015learning}.
We used RDF2Vec\cite{RDF2Vec}, SimpleIE~\cite{SimpleIE}, TransE~\cite{TransE}, KGloVe~\cite{cochez2017global}, CrossE~\cite{CrossE}, and PBG~\cite{lerer2019pytorch}  for the KG embeddings.
These representations represent the neighborhood of a node as well as the kind of relations that exist to the neighboring nodes.
Since these methods do not incorporate literal information into the embedding, literals are removed from the KG.

\emph{RDF2Vec} works by first generating a corpus of text by performing uniform random walks starting from each entity in the graph~\cite{cochez2017biased}.
Then, the corpus $\mathcal{C}$ of edge-labeled random walks are used as the input for learning embeddings of each node using the skip-gram~(SG) word2vec~\cite{mikolov2013efficient} model\footnote{Literature~\cite{cochez2017biased,celebi2018evaluation} have observed better performance using SG than the CBOW model}.
From a given a sequence of drug facts $(w_1,w_2,...,w_n)\in \mathcal{C}$, the SG model aims to maximize the average log probability $L_p$ (see \cref{eq:log}) according to the context within the fixed-size window, in which c represents a context. 

\begin{equation}
    L_p= \frac{1}{N} \sum_{n=1}^{N} \sum_{-c \leq j \leq c, j \neq 0} \log p\left(w_{n+j} | w_{n}\right)
    \label{eq:log}
\end{equation}

To define $p\left(w_{n+j} | w_{n}\right)$, we use negative sampling by replacing $\log p\left(w_{O} | w_{I}\right)$ with a function to discriminate target words $(w_o)$ from a noise distribution $P_n(w)$ drawing $k$ words from $P_n(w)$:

\begin{equation}
    \log \sigma\left(v_{w_{O}}^{\prime \top} v_{w_{I}}\right)+\sum_{i=1}^{k} \mathbb{E}_{w_{i}} \sim_{P_{n}(w)}\left[\log \sigma\left(-v_{w_{i}}^{\prime \top} v_{w_{I}}\right)\right]
    \label{eq:kutta}
\end{equation}

The embedding of a concept $s$ occurring in corpus $\mathcal{C}$ is the vector $v_s$ in \cref{eq:kutta} derived by maximizing \cref{eq:log}.
Besides RDF2Vec, we also trained \emph{TransE} embeddings as a representative of the translation-based KG embedding methods.
Here, every entity and relation is embedded as a low-dimensional vector, where the relations are represented as the translation from the head entity to tail entity.
The \emph{CrossE} embedding method, which explicitly simulates crossover interactions, is also used.
Although both general embeddings for each entity and relation and multiple triple specific embeddings can be generated using CrossE, we used only the general embeddings. 
The \emph{SimplE} embedding method, which allows two embeddings of each entity to be learned independently, is also employed.
The embeddings learned through SimplE are interpretable, which help to incorporate drug-related background knowledge into the embeddings.
A reported advantage of SimplE is that it 
outperforms tensor factorization techniques, especially for link prediction problems. 

Learning the representations in a KG relies on contrasting positive instances with negative ones. 
However, KGs typically include only positive relation instances\cite{socher2013reasoning}. 
A solution to this issue is using implicit negative evidence in which instances that have not been observed in the KGs are considered negative. 
Kotnis and Nastase et al.~\cite{kotnis2017analysis} employed several negative sampling approaches and observed the impact on the learned embeddings. They found that the ``corrupting positives" method is leading to the best result in a link prediction task. 
This corruption produces negative instances that are closer to the positive ones than those produced through random sampling. 
Also several methods which we used in this work (\emph{TransE}, \emph{CrossE}, \emph{SimplE}, and \emph{RDF2Vec}) generate negative instances by corrupting positive samples. 


\emph{KGloVe}~\cite{cochez2017global} has some similarity with RDF2Vec, but uses a different technique to identify global patterns for creating vector space embeddings. 
First, a co-occurrence matrix is created by personalized computing PageRank for each node.
Then, this process is repeated for the graph with all edges reversed.
These two matrices are summed together and normalized.
Finally, this matrix is used as an input to the GloVe word embedding algorithm.
In this work, we have only used unbiased walks.
As a last model, we train ComplEx~\cite{trouillon2016complex} using the \emph{PBG} implementation~\cite{lerer2019pytorch} because our integrated KG contains many triples. 
Technically, ComplEx uses only the Hermitian dot product for creating embeddings\footnote{Complex counterpart of the standard dot product between real vectors}. 
The overall embedding method using ComplEx is arguably simpler but since the composition of complex embeddings can handle a large variety of binary relations (among them symmetric and antisymmetric ones) research has exposed that complex embeddings can outperform several other models. 

We trained the ComplEx embedding model of PBG to i) create the embeddings faster, scalable for large graphs, and parallelize the training, and ii) to observe if the embeddings generated by this model are useful for predicting DDIs.
Given these dense representations, we can now feed the information from the graph into our machine learning models.
To represent the feature vector of a drug pair, we concatenate the embedding vectors of each drug in the pair in which the negative samples are generated by corrupting positive edges by sampling either a new source or a destination for each existing edge~\cite{lerer2019pytorch}.

\subsection{Network construction}
\begin{figure*}
	\centering
	\includegraphics[width=\textwidth]{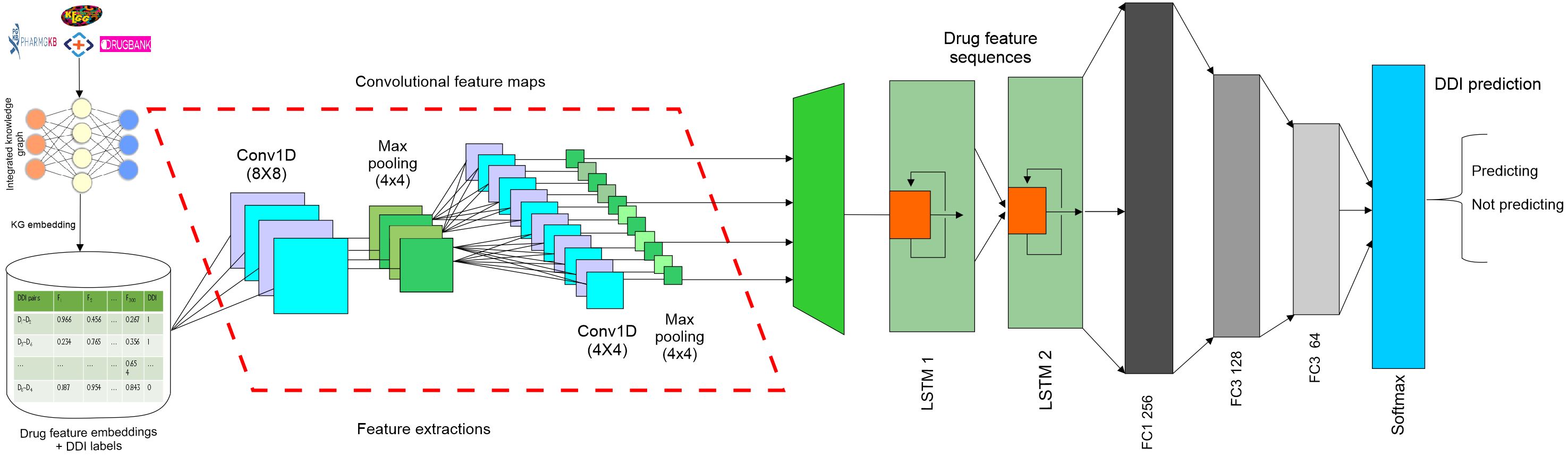}	
	\caption{A schematic representation of the \texttt{Conv-LSTM} network, which starts from taking input into an n-dimensional embedding space and passing to both CNN and LSTM layers before getting the vector representation of the most important features to fed through dense, dropout, Gaussian noise, and Softmax layers for predicting possible drug-drug interactions.}	
	\label{fig:network} 
\end{figure*}

We train various baseline ML models, which we will use as baselines later.
Here, we describe the more complex neural network architecture which gave the best results we obtained.
We construct a so-called \texttt{Conv-LSTM} network~\cite{Conv_LSTM1} by combining both CNN and LSTM layers as shown in \cref{fig:network}.
While CNN uses convolutional filters to capture local relationship values in drug features, the LSTM network can carry overall relationships from the features extracted by CNN. 
The \texttt{Conv-LSTM} has shown good performance on diverse prediction tasks such as hate speech detection from text~\cite{zhang2018hate}, for precipitation nowcasting~\cite{Conv_LSTM1}, and for monocular depth prediction~\cite{cs2018depthnet}. 
Consequently, it has been able to capture the most significant drug features in our case.

We extended the \texttt{Conv-LSTM} network proposed in literature ~\cite{Conv_LSTM1} in which each input $\mathcal{X}_{1},\mathcal{X}_{2},...,\mathcal{X}_{t}$, cell outputs $\mathcal{C}_{1},\mathcal{C}_{2},...,\mathcal{C}_{t}$, hidden states $\mathcal{H}_{1},\mathcal{H}_{2}....,\mathcal{H}_{t}$, and gates $i_t$,$f_t$,$o_t$ of the network are 2D tensors whose dimensions are spatial dimensions of the drug features.
\texttt{Conv-LSTM} determines the future state of a certain cell in the input hyperspace by the inputs and past states of its local neighbors.
This is achieved by using a conv operator in the state-to-state and input-to-state transitions as represented as follows~\cite{Conv_LSTM1}:

\begin{align} 
i_{t} &=\sigma\left(W_{x i} * \mathcal{X}_{t}+W_{h i} * \mathcal{H}_{t-1}+W_{c i} \circ \mathcal{C}_{t-1}+b_{i}\right) \\ 
f_{t} &=\sigma\left(W_{x f} * \mathcal{X}_{t}+W_{h f} * \mathcal{H}_{t-1}+W_{c f} \circ \mathcal{C}_{t-1}+b_{f}\right) \\ \mathcal{C}_{t} &=f_{t} \circ \mathcal{C}_{t-1}+i_{t} \circ \tanh \left(W_{x c} * \mathcal{X}_{t}+W_{h c} * \mathcal{H}_{t-1}+b_{c}\right) \\ 
o_{t} &=\sigma\left(W_{x o} * \mathcal{X}_{t}+W_{h o} * \mathcal{H}_{t-1}+W_{c o} \circ \mathcal{C}_{t}+b_{o}\right) \\ \mathcal{H}_{t} &=o_{t} \circ \tanh \left(\mathcal{C}_{t}\right) 
\end{align} 

In the above equations, $*$ denotes the conv operator, and $\circ$ is the entrywise multiplication of two matrices of same dimensions.
The second LSTM layer emits an output `$\mathcal{H}$,' which is then reshaped~(i.e., flatten) into a feature sequence and fed into fully-connected layers to predict the DDIs at the next step and as an input at the next time step.
The first layer is the embedding layer, which maps a drug sample as a 'sequence' into a real vector domain.
Then the embedding representation with a shape of 100x300 is fed into a 1D convolutional layer, which has 100 filters and a kernel-size of 4. 

The output of each conv layer is then passed to the dropout layer to regularize learning to avoid overfitting.
Intuitively, these can be thought of as forcing the classifier not to rely on any trivial individual drug features. 
The conv layer convolves the input feature space into a 100x100 representation, which is further down-sampled by the 1D max pooling layer~(MPL) having a pool size of 4 along the embedding dimension, producing an output of shape 25x100. Where each of the 25 dimensions can be considered as an 'extracted feature.' 
The MPL flattens the output space by taking the highest value in each timestep dimension, which produces a 1x100 vector containing drug features that are highly indicative of interest.
Contrarily, LSTM layer treats flattened feature vector's dimension as timesteps and outputs 100 hidden units per timestep.
Then using a global MPL, the most influential features are fed into a fully-connected layer after passing through another dropout layer and finally to a softmax layer which generates the probability distribution over the classes.
Additionally, we introduce Gaussian noise~\cite{xie2012image} into each conv, LSTM, and dense layer to improve the model generalization.

\subsection{Network training}
Since all the classifiers need both negative and positive samples for the link prediction problem, previous studies have randomly chosen negative samples from unknown interactions~\cite{confusion4,sridhar2016probabilistic}.
However, setting all the unknown interactions as negative samples creates a data imbalance issue.
Consequently, performance metrics, such as AUPR and F1-score, get influenced~\cite{celebi2018evaluation}.
Other research has tackled this issue through random undersampling from the unknown interactions at a ratio corresponding to the positive set~\cite{cheng2014machine}, or inferring negatives by unsupervised clustering analysis~\cite{hameed2017positive}. 

The open source implementations of PBG\footnote{\url{https://github.com/facebookresearch/PyTorch-BigGraph}},  CrossE\footnote{\url{https://github.com/wencolani/CrossE}}, TransE\footnote{\url{https://github.com/xjdwrj/TransE-Pytorch}}, and SimpleIE\footnote{\url{https://github.com/baharefatemi/SimplE}} were used for the KG embedding with the default parameters provided.
On the other hand, the modified version of KGloVe\footnote{\url{https://github.com/miselico/globalRDFEmbeddingsISWC}} is used, which converged at $400^{th}$ iteration.
While RDF2Vec was trained using skip-gram by setting window size = 5 with graph walk at depth 5 and 500 walks per entity. 
Each embedding methods is employed by setting the dimension of the feature vector to 300 by varying negative samples.
By filtering out the drugs that have no calculated feature vector, we were able to extract the features for 12,439 drugs out of 12,664 drugs.
The embeddings generated by RDF2Vec, TransE, PBG, KGloVe, CrossE, and SimpleIE are then used to train the Conv-LSTM network for the link prediction similar to~\cite{btx275}, in which we aim to estimate the probability that a relation or link with label $l$ exists between vertices $v_1$ and $v_2$ given their vector representation, $V(v_1)$ and $V(v_2): p(v_1,v_2,l) \in E |\left\langle V\left(v_1\right), V \left(v_{2}\right)\right\rangle$. 

\begin{equation}
    L=\sum_{i, j \in Y}-y_{i, j} \log \hat{y}_{i, j}-\left(1-y_{i, j}\right) \log \left(1-\hat{y}_{i, j}\right)
    \label{eq:cel}
\end{equation}

First-order gradient-based optimization techniques Adam, AdaGrad, RMSprop, and AdaMax with varying learning rates and different batch sizes are used to learn model parameters, which tries to optimize the \texttt{binary cross-entropy} loss~\cref{eq:cel}. The hyperparameters optimization is done based on random search and cross-validation in which the model is trained on a batch size of 128 wherein each of 5 runs 70\% of the data is used for the training, 30\% for evaluating the network, and 10\% from the training set is randomly used for the validation.

\section{Experiments}
\label{results}

The evaluation code\footnote{Source code: \url{https://github.com/rezacsedu/DDI-prediction-KG-embeddings-Conv-LSTM}} was written in Python.
The software stack consists of Scikit-learn, PyTorch, and Keras with the TensorFlow backend.
The network training is carried out on an Nvidia GTX 1080i GPU with CUDA and cuDNN enabled. 
We also trained LR, KNN, NB, SVM, RF, and GBT as ML baseline models.
Similar to Conv-LSTM network, we perform the hyperparameter optimization for these classifiers through a random search and 5-fold cross-validation tests.
For the experiment, 80\% of the data is used for the training using 5-fold cross-validation and evaluate the optimized model on 20\% held-out data in which the best hyperparameters were produced through a random search. 
Although AUC score is used commonly as a performance metric in previous studies, literature has emphasized that it might not be sufficiently accurate for imbalanced data~\cite{celebi2018evaluation,kastrin2018predicting}. 
Therefore, we used the area under the precision-recall curve~(AUPR), and Matthias correlation coefficient~(\texttt{MCC}) along with the AUC and F1-score to measures the performance of the classifiers. Finally, we use the model averaging ensemble~(MAE) of top-3 models to report the final prediction.
 
\subsection{\textbf{Analysis of DDIs predictions}} 
\label{sec:performance-sentiment-analysis}

\begin{table}
	\caption{Prediction comparison across embedding methods. * signifies the embedding method giving best accuracy.}
	\label{table:all}
	\small
	\begin{tabular}{clrrr}
		\toprule
		\textbf{Embedding} & \textbf{Model} & \textbf{AUPR} & \textbf{F1-score} & \textbf{MCC}\\ \hline
		\multirow{7}{*}{ComplEx*} & LR & 0.74 & 0.72 & 0.53 \\
		& NB & 0.73 & 0.70 & 0.55 \\
		& SVM & 0.80 & 0.81 & 0.69 \\
		& KNN & 0.69 & 0.71 & 0.51 \\
		& GBT & 0.88 & 0.85 & 0.72\\
		& RF & 0.91 & 0.92 & 0.75\\
		& \textbf{Conv-LSTM}  & 0.93 & 0.91 & 0.79 \\
		\cmidrule(l){2-5}
		& \textbf{MAE} & 0.94 & 0.92 & 0.80 \\ \midrule
		\multirow{7}{*}{SimpleIE} & LR & 0.76 & 0.72 & 0.59 \\
		& NB & 0.73 & 0.72 & 0.55 \\
		& SVM & 0.80 & 0.81 & 0.66 \\
		& KNN & 0.73 & 0.73 & 0.53 \\
		& GBT & 0.87 & 0.86 & 0.74\\
		& RF & 0.89 & 0.88 & 0.77\\
		& Conv-LSTM  & 0.91 & 0.90 & 0.78 \\
		\cmidrule(l){2-5}
		& MAE & 0.92 & 0.91 & 0.79 \\ \midrule
		\multirow{7}{*}{KGloVe} & LR & 0.75 & 0.73 & 0.54 \\
		& NB & 0.72 & 0.71 & 0.53 \\
		& SVM & 0.78 & 0.79 & 0.68 \\
		& KNN & 0.71 & 0.69 & 0.53 \\
		& GBT & 0.87 & 0.85 & 0.71\\
		& RF & 0.89 & 0.86 & 0.73\\
		& Conv-LSTM  & 0.87 & 0.89 & 0.74 \\
		\cmidrule(l){2-5}
		& MAE & 0.89 & 0.90 & 0.75 \\ \midrule
		\multirow{7}{*}{TransE} & LR & 0.72 & 0.71 & 0.57 \\
		& NB & 0.69 & 0.70 & 0.51 \\
		& SVM & 0.75 & 0.74 & 0.64 \\
		& KNN & 0.63 & 0.59 & 0.49 \\
		& GBT & 0.83 & 0.82 & 0.69\\
		& RF & 0.84 & 0.85 & 0.71\\
		& Conv-LSTM  & 0.86 & 0.87 & 0.73 \\
		\cmidrule(l){2-5}
		& MAE & 0.87 & 0.88 & 0.74 \\ \midrule
		\multirow{7}{*}{CrossE} & LR & 0.65 & 0.68 & 0.47 \\
		& NB & 0.70 & 0.71 & 0.50 \\
		& SVM & 0.72 & 0.73 & 0.55 \\
		& KNN & 0.69 & 0.66 & 0.52 \\
		& GBT & 0.81 & 0.82 & 0.65\\
		& RF & 0.82 & 0.83 & 0.66\\
		& Conv-LSTM  & 0.85 & 0.84 & 0.67 \\
		\cmidrule(l){2-5}
		& MAE & 0.86 & 0.85 & 0.69 \\ \midrule
		\multirow{7}{*}{RDF2Vec} & LR & 0.65 & 0.68 & 0.47 \\
		& NB & 0.69 & 0.70 & 0.48 \\
		& SVM & 0.71 & 0.72 & 0.52 \\
		& KNN & 0.67 & 0.64 & 0.50 \\
		& GBT & 0.79 & 0.80 & 0.61\\
		& RF & 0.80 & 0.81 & 0.62\\
		& Conv-LSTM  & 0.83 & 0.82 & 0.67 \\
		\cmidrule(l){2-5}
		& MAE & 0.84 & 0.85 & 0.69 \\ \bottomrule
	\end{tabular}
\end{table}

\Cref{table:all} summarizes the results of the prediction task based on different embedding methods. 
A general observation is that the \texttt{Conv-LSTM} model outperformed all baseline models, in the best case resulting in an AUPR of 0.93.
Also, overall LR, NB, KNN, and SVM models performed worst.
Although these algorithms are intrinsically simple, have low variance, and less prone to over-fitting, feature selector based on these may be discriminating drug-related features very aggressively, which forces these classifiers to lose some useful drug features which result in worse performance. 
Among the tree-based classifiers, RF performs the best, showing an F1-score of 0.91, which is the best among the ML baselines.
The model averaging ensemble of top-3 models~(i.e., GBT, RF, and Conv-LSTM) boosts the performance by 1.5\%compared to the best \texttt{Conv-LSTM} model 
in terms of F1 score. 

\begin{figure}
	\centering
	\includegraphics[width=0.95\linewidth]{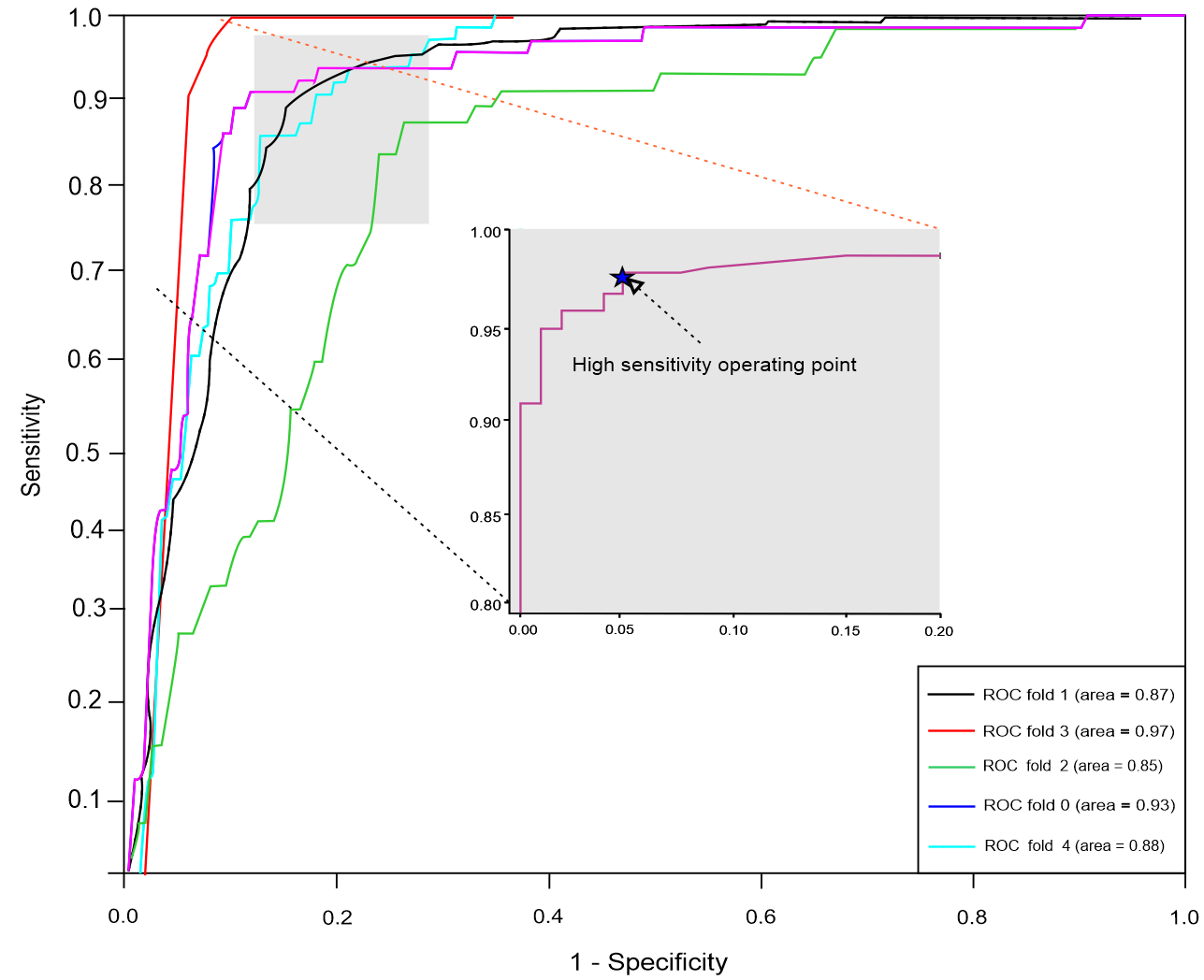}
	\caption{ROC curves of cross-validated Conv-LSTM model} 
	\label{fig:roc}
\end{figure}

Interestingly, the \texttt{MCC} scores by all classifiers also show the prediction was strongly correlated with the ground truth (measured using a Pearson product-moment correlation coefficient we obtained 0.70), probably because the embeddings generated by the embedding methods are learnable quality drug features. 
The AUC score generated by the \texttt{Conv-LSTM} network is found to be the highest, which is at least 3\% better than the second-best score by the RF classifier, where the LR classifier performed the worst.
The ROC curve in \cref{fig:roc} shows consistent AUC scores across the folds, which signifies that the predictions are much better than random guessing. 

\subsection{Comparison of graph embedding methods}
\begin{figure}
	\centering
	\includegraphics[width=\linewidth]{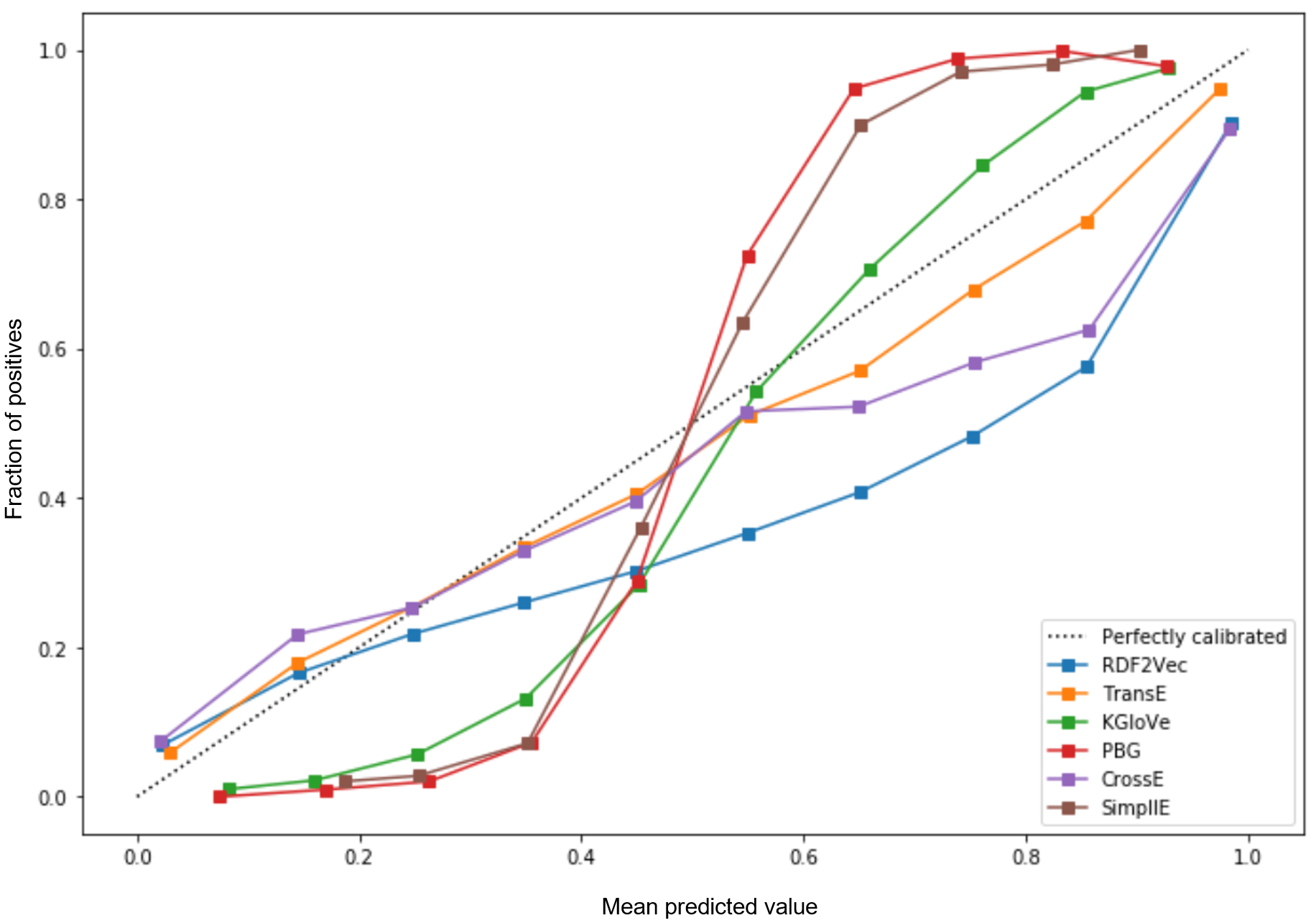}	
	\caption{Calibrating Conv-LSTM with embedding methods}	
	\label{fig:cal}
\end{figure}

As seen in \cref{table:all}, the classifiers work better with drug features generated by the ComplEx, SimpleIE, and KGloVe methods.
In particular, the GBT, RF, and Conv-LSTM classifiers perform consistently best on the embeddings generated by ComplEx in terms of F1, MCC, and AUPR scores.
In contrast, using features generated by the RDF2VEc method, we experience the worst DDIs prediction accuracy.
This is different from earlier work~\cite{celebi2018evaluation}, where the best results were obtained with RDF2Vec with uniform weighting setting. 
We suspect that the classifiers did not benefit much from more training samples in our case.  
Therefore, we validate this by calibrating the best performing Conv-LSTM classifier against different embedding methods for which the output probability of the classifier can be directly interpreted as a confidence level in terms of `fraction of positives'; the result is illustrated in~\cref{fig:cal}. 
As seen, the Conv-LSTM classifier gave a probability value between 0.82 to 0.93, which means 93\% predictions belong to true positive predictions generated by the embeddings from PBG. 

\subsection{Effects of number of drug samples}
\begin{figure*}
	
	\centering
	\begin{subfigure}{0.48\linewidth}
		\centering
		\includegraphics[width=\linewidth]{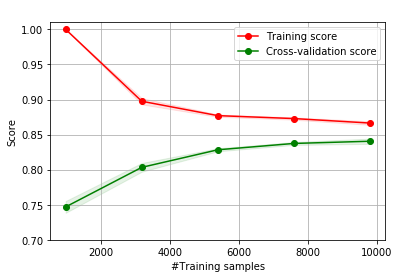}
		\caption{SVM}
		\label{fig9a}
	\end{subfigure}
	\begin{subfigure}{0.48\linewidth}
		\centering
		\includegraphics[width=\linewidth]{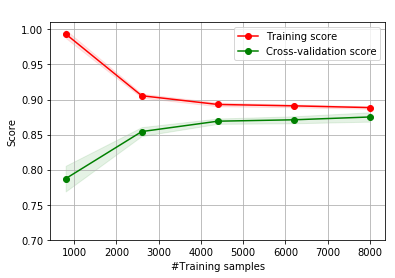}
		\caption{GBT}
		\label{fig9b}
	\end{subfigure}
	\begin{subfigure}{0.48\linewidth}
		\centering
		\includegraphics[width=\linewidth]{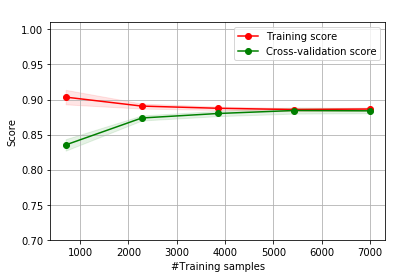}
		\caption{RF}
		\label{fig9c}
	\end{subfigure}
	\begin{subfigure}{0.48\linewidth}
		\centering
		\includegraphics[width=\linewidth]{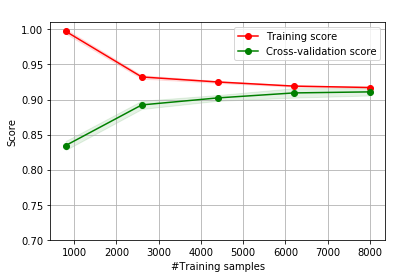}
		\caption{Conv-LSTM}
		\label{fig9d}
	\end{subfigure}
	\caption{Learning curves showing the validation and training scores of top-3 and SVM classifier} 
	\label{fig:lc}
	
\end{figure*}

To understand the effects of having more training samples, and to understand whether our classifiers suffer more from variance errors or bias errors, we observed the learning curves of top-3 classifiers (i.e., RF, GBT, and Conv-LSTM) and SVM (a linear model) for varying numbers of training samples.
As shown in \cref{fig:lc}, for SVM the validation and training scores converge to a low value with increasing size of the training set.
Consequently, SVM did not benefit much from more training samples. 
However, RF and GBT are tree-based ensemble methods, and the Conv-LSTM network can learn more complex concepts from the drug features.
This results in a lower bias, which can be observed from higher training scores than the validation scores for the maximum number of drug samples, i.e., adding more training samples does increase generalization.

\subsection{Influence of negative samples}
Inspired by~\cite{trouillon2016complex}, we investigated the influence of the number of negatives per positive training sample, which we call $\sigma$, for ComplEx.
As we already varied $\sigma$ per training sample and set it to 15, we further varied again the $\sigma$ in [5, 10, 20, 25] and collected the embeddings for each setting again.
Then we observed if the Conv-LSTM network performs better with the larger $\sigma$. Generating more negatives samples moderately improves the results.
In particular, with 20 negative triples, we observed about 1\% accuracy boost in terms of AUPR. 
Embedding training also converged slightly faster.
Further increasing $\sigma$ to 25 results in a drop in AUPR. 

\subsection{Comparison with state-of-the-art}
Since our approach of data collection and preparation are different from other approaches and we have more samples, a one-to-one comparison was not viable --especially with Tiresias framework~\cite{abdelaziz2017large}, PRD\cite{wang2017predicting}, and INDI~\cite{gottlieb2012indi}. 
Kastrin et al.~\cite{kastrin2018predicting} used data from multiple sources, but evaluated and inferred unknown DDIs from TWOSIDES only.
With that dataset, they achieved the best AUPR score of 0.93 using RF and GBT classifiers. 

The Tiresias framework, which uses both pharmacological similarities from embedding features, has reported an F1-score of 0.85 and AUPR of 0.92. Their pharmacological similarity features are equivalent to INDI, which also uses the DrugBank v4.0 dataset.
INDI evaluated the performance of DDI prediction on a total of 37,212 true DDIs.
They obtained AUC scores of 0.93 and an F1-score of 0.89, omitting the interaction type (i.e., PD or PK) using a 10-fold cross validation setting.
Remzi et al.~\cite{celebi2018evaluation} observed an F1-score of 0.867 and AUPR of 0.918 using DrugBank v4.0 dataset. 

With our approach, evaluations against several baseline models yield an AUPR of up to 0.94, an F1-score of 0.92, and an MCC of 0.80 during 5-fold cross-validation tests.
This signifies that a KG-based approach using multiple data sources is comparable to current state-of-the-art methods.
To show the benefit of using a more robust classifier, we further trained the \texttt{Conv-LSTM} network with DrugBank v4.0 dataset as shown in~\cref{fig:workflow}. During a 5-fold cross-validation test, we observed slightly better accuracy, 
namely an F1-score of 0.895 and AUPR of 0.926, which outperforms~\cite{celebi2018evaluation}.

\section{Conclusion and outlook}
\label{conclusion}

Adverse drug reactions are very dangerous and lead to a significant number of the hospital (re-)admissions and even deaths.
Many of these reactions are due to drug-drug interactions.
Preferably, all drug-drug interactions should be known upfront to ensure that preventable cases do not occur.
However, it is not feasible to investigate all possible interactions, and hence approaches able to predict possible interactions are investigated.
In this paper, we proposed the use of knowledge graphs to integrate drug-related data from different sources.
This way, we have integrated background knowledge about drugs, diseases, pathways, proteins, enzymes, chemical structures, etc.
Since this background data is in a format which cannot be used as a direct input for typical classifiers, we applied several node embedding techniques to create a dense vector representation for each node in the KG.
These representations are then fed into various traditional ML classifiers and a specifically designed neural network architecture based on a convolutional-LSTM.

Our core observations are that 
\begin{inparaenum}[i)]
\item We could outperform the baseline classifiers, as well as earlier state-of-the-art models, consistently with our proposed architecture. 
We obtained up to 0.94, 0.92, 0.80 for AUPR, F1-score, and MCC, respectively, during 5-fold cross-validation tests showing high confidence at predicting potential DDIs.
\item From the embedding models we used, the PBG model did perform best, but also SimpleE and KGloVe gave reasonable results.
\end{inparaenum}

One limitation of our approach is the inability to provide explanations for the predicted DDIs.
The embedding creates latent features, which are like a black-box model.
As future research directions, we see 
\begin{inparaenum}[i)]
\item the possibility to include even more data to the background. For example, NDFRT, SemMedDB, and SIDE; also a large ablation study to measure the influence of each of these additions would be useful,
\item also including explicit information about negative drug-drug interaction, as well as a prediction of the interaction type, 
\item providing explanations for the interactions, 
\item further investigation of other models to perform predictions on graphs, and
\item interaction with other (non-drug) substances like food.
\end{inparaenum}

\newpage

\bibliographystyle{ACM-Reference-Format}
\bibliography{sample-base}

\end{document}